\documentclass[letterpaper, 10 pt, conference]{ieeeconf}  % Comment this line out if you need a4paper

\IEEEoverridecommandlockouts                              % This command is only needed if 
\usepackage{graphicx}
\usepackage[dvipsnames]{xcolor}

\title{\LARGE \bf
 Can I Pet Your Robot? Incorporating Capacitive Touch\\ Sensing into a Soft Socially Assistive Robot Platform
}

\author{Amy O'Connell$^{1}$, Bailey Cislowski$^{1}$, Heather Culbertson$^{1}$, and Maja Matari\'c$^{1}$% <-this % stops a space
\thanks{This research was supported by the National Science Foundation Grant NSF IIS-1925083.}% <-this % stops a space
\thanks{$^{1}$Amy O'Connell, Bailey Cislowski, Heather Culbertson, and Maja Matari\'c are with the Department of Computer Science, University of Southern California, Los Angeles, CA, USA {\tt\small amy.dell@usc.edu, bcislows@usc.edu, hculbert@usc.edu, mataric@usc.edu}}}

\begin{document}

\maketitle
\thispagestyle{empty}
\pagestyle{empty}

\begin{abstract}

This work presents a method of incorporating low-cost capacitive tactile sensors on a soft socially assistive robot platform. By embedding conductive thread into the robot's crocheted exterior, we formed a set of low-cost, flexible capacitive tactile sensors that do not disrupt the robot's soft, zoomorphic embodiment. We evaluated the sensors' performance through a user study (N=20) and found that the sensors reliably detected user touch events and localized touch inputs to one of three regions on the robot's exterior.

\end{abstract}

\section{Motivation}

Touch plays an important role in human-human and human-animal social interactions, but it is not often utilized in social interactions between humans and robots. Most commercially available platforms for robotics research do not incorporate tactile sensing, and those that do cannot be easily modified to fit the unique needs of a particular interaction, context, or user. In this work, we propose and evaluate a method of incorporating low-cost capacitive tactile sensors onto the Blossom open-source soft robot platform~\cite{suguitan2019blossom} to enable low-cost, customizable tactile sensing in human-robot interactions. We chose Blossom, a zoomorphic tabletop robot with a soft, crocheted exterior, because it is a low-cost, expressive, highly customizable platform for human-robot interaction (HRI) research which has been for various applications in social and socially assistive contexts. Blossom's original design did not feature tactile sensing. 

Several soft zoomorphic robots utilize tactile sensing for physical interactions with users, such as Paro \cite{shibata2001mental}, Huggable \cite{stiehl2005design}, and the Haptic Creature \cite{yohanan2012role}. 
The tactile sensors built into each of these platforms are highly specific to the robot's embodiment and can not be easily reconfigured for different types of interactions.

\begin{figure}[t!]
  \centering
  \framebox{\parbox{2.5in}{
  \centering
  \includegraphics[width=2.5in]{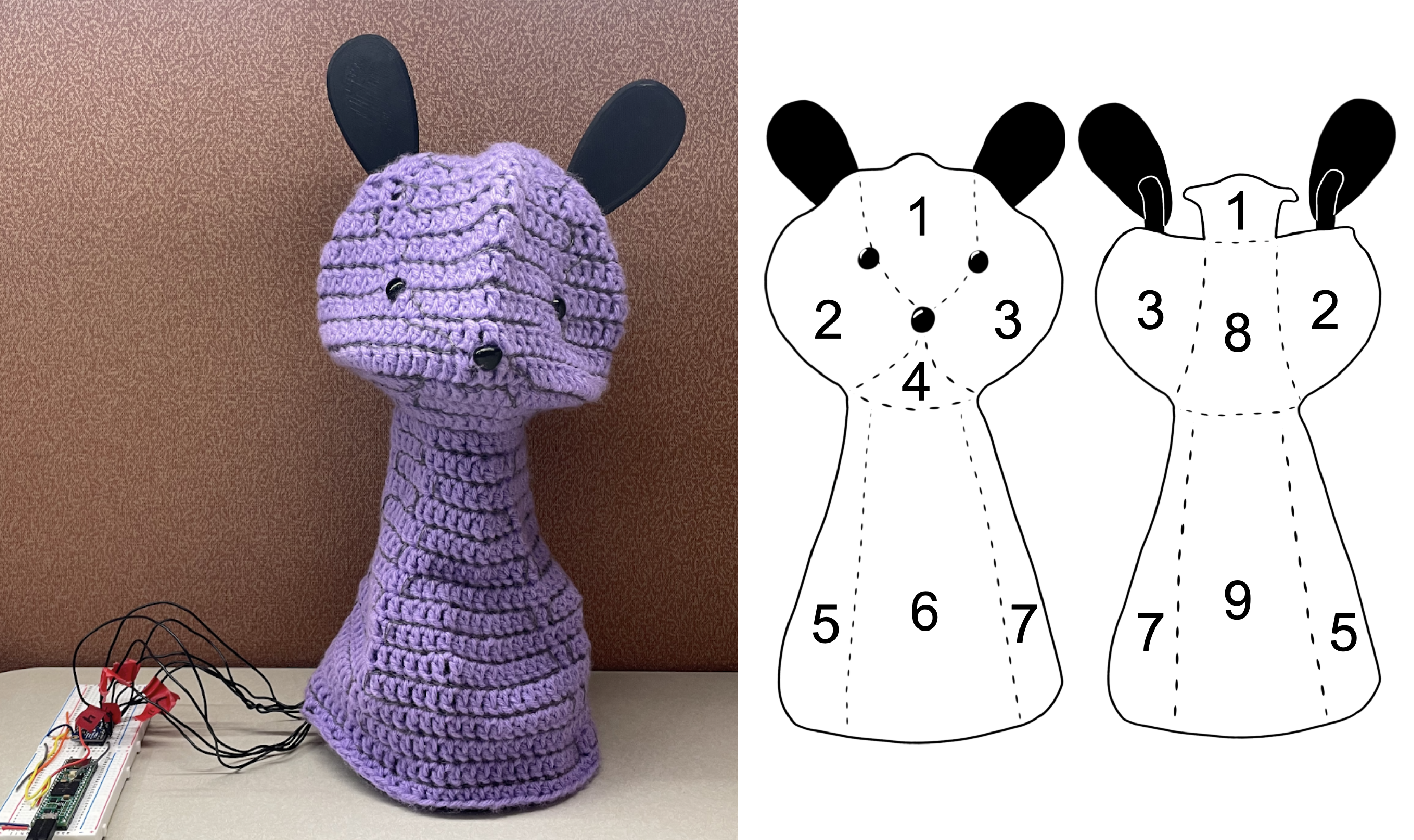}}}
  \caption{The robot's exterior features nine soft capacitive touch sensors}
  \label{full_skin}
\end{figure}

\section{Device Design}

We developed a low-cost, customizable system of tactile sensors for the Blossom open-source robot platform. 
The system satisfies the following design goals:\\
\textbf{System} Equip Blossom's soft, crocheted exterior with sensors to detect users' touch. Sensors should be constructed from low-cost, widely available materials and hardware, making them accessible for projects with limited resources.\\
\textbf{Performance} Sensors should be able to reliably detect and localize users' touch.\\
\textbf{Flexibility and Cohesiveness} Sensors should be easy to reconfigure for different interaction contexts without disrupting Blossom's soft, handcrafted embodiment.

\subsection{Physical Device Design}

\begin{figure}[t]
  \centering
  \framebox{\parbox{1.69in}{
  \centering
  \includegraphics[width=1.69in]{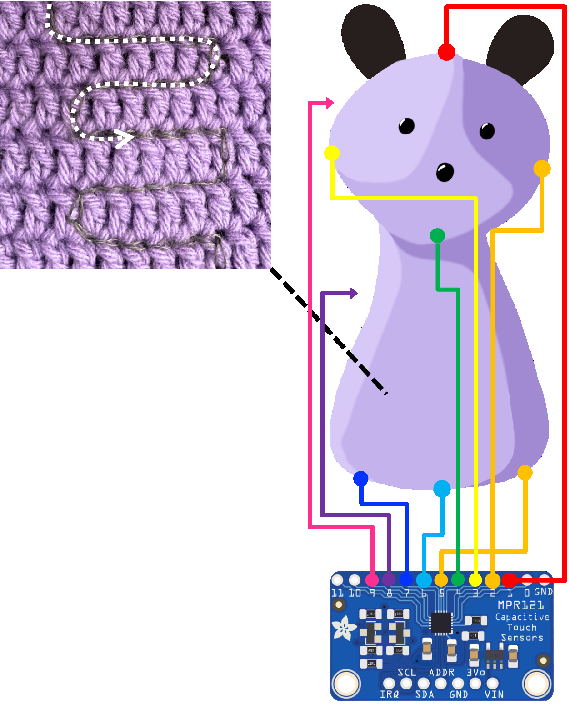}}}
  \caption{Nine conductive thread sensors were connected to a capacitive touch sensor breakout board (Adafruit MPR121) and Teensy microcontroller}
  \label{system_diagram}
  % \vspace{-4pt}
\end{figure}

We embedded conductive thread into the crocheted exterior to enable Blossom to detect touch from the user. The exterior was divided into nine regions determined by landmark features on the robot's face and body (e.g., buttons, flaps, and seams).
We used the surface crochet technique, pictured in Fig. \ref{system_diagram}, to embed the sensors into the robot's crocheted exterior. The threads in each of the nine regions were attached to individual lengths of 18-gauge insulated solid copper wire, and routed to a single posterior point on the bottom edge of the robot's exterior. The wires were connected to a capacitive touch sensor breakout board (Adafruit MPR121) itself connected to a microcontroller (Teensy 4.1).
The complete system is shown in Fig. \ref{full_skin}. In a timed trial, an experienced crocheter installed one 3"x4" sensor (8 rows of 13 double crochets) in 10 minutes and uninstalled the sensor within 30 seconds, meaning the sensors can be reinstalled in different configurations with relative ease.

The microcontroller was connected via USB to a laptop and transmitted filtered and baseline outputs for each sensor to the computer via serial communication at a rate of 9.1 Hz.

\subsection{Software Design}
The MPR121 detected touch and release events by comparing the electrode-filtered data to a baseline value, where the difference between the baseline value and filtered value represented an immediate capacitance change associated with a touch event. Individual sensors were classified as "touched" during a gesture if the difference between the baseline value $B$ and filtered value $F$ exceeded a threshold $T$:

$$
|B - F| \geq T \eqno{(1)}
$$

The threshold $T$ determines the sensitivity of the tactile sensors, so lower $T$ values detect more touch events but may also cause more false positive touch detections. 

\section{EVALUATION}

We conducted a small user study (N=20) to evaluate the sensors' ability to detect users' touch for different types of gestures. The study participants were recruited from a convenience population of university students. They identified as: 7 female, 11 male, 2 non-binary; 10 Asian, 1 Black,  2 Hispanic Latino, 7 white; their ages ranged from 18 to 26 (M=22.35, SD=2.39). Participants were not compensated for completing the study.

Participants were verbally asked to perform five touch gestures on the robot: Contact, Stroke, Pat, Scratch, and Poke. They were instructed to perform the gestures in order on three sensorized regions: the right side of the robot's trunk, the right side of the robot's head, and the top of the robot's head. We chose to evaluate only a subset of the sensors because several sensors had nearly identical construction (e.g., right and left cheek, right and left side of the trunk). The three sensors we chose to evaluate had different underlying mechanical structures (rigid components, flexible joints, and empty space) that could affect their performance.  

Filtered and baseline values for each sensor were logged to a CSV file. Each frame of sensor values was labeled with the gesture and location. After all the data were collected, we classified each gesture as \textit{touched} or \textit{not touched} using Eq. 1 with $T=10$ (determined in empirical testing). 

\section{RESULTS AND DISCUSSION}

Table \ref{results_table} shows the percentage of participants for whom the corresponding touch sensor correctly detected each gesture in each location. Of the three sensors, the top of the head performed the best, detecting 96\% of all gestures performed in that area. The left trunk sensor had a similar performance, detecting 94\% of all gestures. The left cheek sensor performed considerably worse than the other two sensors, detecting only 60\% of all touch events on the robot's cheek.

\begin{table}[t]
\caption{Percent of gestures detected as touch events}
\label{table_example}
\vspace{-0.15in}
\begin{center}
\begin{tabular}{|c||c|c|c|c|c|}
\hline
&contact&stroke&pat&scratch&poke\\
\hline
Trunk&100\%&90\%&100\%&85\%&95\%\\
\hline
Cheek&70\%&60\%&50\%&55\%&65\%\\
\hline
Top of Head&100\%&100\%&100\%&95\%&85\%\\
\hline
\end{tabular}
\end{center}
\label{results_table}
% \vspace{-10pt}
\end{table}

Several factors may have led to the observed difference in performance. First, the robot's construction affected how it responded to gestures in different areas. Blossom's rigid frame features a head platform that is suspended on elastic bands. When the head is pressed downward from above, the elastic bands stretch downward in unison, exerting an opposing force that the user perceives as resistance). However, when a force is exerted on the head from the side, the head platform swings in the direction of the force with little resistance, giving the robot a "bobblehead" feeling. Although the sensors did not measure pressure intensity, the opposing force from the top of the head may have caused more contact between the sensor and the user's skin, leading to more pronounced signal changes and more consistent touch detection than gestures performed on the cheek. 

The social nature of touch gestures may have also played a role in how participants performed gestures in each location. For example, participants may have felt it was more natural to pat the top of the robot's head than the robot's cheek from the side. One participant commented that the gestures performed on the cheek felt unwelcome due to the movement of the suspended head platform: "It feels like [Blossom]'s recoiling away from me like she doesn't like it. I'm worried I'll knock her over." Future work will consider accounting for the observed differences in sensor performance by setting individual detection thresholds for each sensor. 

The validated touch sensing exterior will enable HRI researchers to design rewarding physical human-robot interactions. Future work will explore responses to users' touch that encourage adherence to long-term interventions with socially assistive robots. Enabling socially assistive robots to localize touch gestures from tactile sensor data will also allow for more varied and appropriate responses to users' touch. 

Finally, while the presented system of soft tactile sensors was designed for physical human-robot interactions, the potential applications extend beyond HRI. The soft, crocheted sensors could be incorporated into soft interfaces, actuators, or in handcrafted wearables.

\addtolength{\textheight}{-12cm}

\bibliographystyle{IEEEtran}
\bibliography{IEEEabrv, root}

\end{document}